\documentclass[letterpaper, 10 pt, conference]{ieeeconf}  % Comment this line out if you need a4paper

\IEEEoverridecommandlockouts                              % This command is only needed if
\usepackage{comment}
\usepackage{mathrsfs}
\usepackage{amsfonts}
\usepackage{psfig}
\usepackage{graphics} % for pdf, bitmapped graphics files
\usepackage{epsfig} % for postscript graphics files
\usepackage{times} % assumes new font selection scheme installed
\usepackage{amsmath} % assumes amsmath package installed
\usepackage{amssymb}  % assumes amsmath package installed
\usepackage{multicol}
\usepackage{upgreek}
\usepackage{bm}
\usepackage{subfigure}
\usepackage{cite}
\usepackage{subeqnarray}
\usepackage{cases}
\usepackage{multirow}
\usepackage{graphics} % for pdf, bitmapped graphics files
\usepackage{epsfig} % for postscript graphics files
\usepackage{mathptmx} % assumes new font selection scheme installed
\usepackage{times} % assumes new font selection scheme installed
\usepackage{booktabs}
\usepackage{float}
\usepackage{sansmath}
\usepackage{color}
\usepackage{relsize}
\usepackage{comment}
\usepackage{textcomp}

\newenvironment{david}{\par\color{black}}{\par}
\title{\LARGE \bf
Roboat II: A Novel Autonomous Surface Vessel for Urban Environments
}
\author{
~Wei Wang, Tixiao Shan, Pietro Leoni, David Fern\'andez-Guti\'errez, Drew Meyers, Carlo Ratti and Daniela Rus
\thanks{This work was supported by grant from the Amsterdam Institute for Advanced Metropolitan Solutions (AMS) in Netherlands.}
\thanks{W. Wang, T. Shan, P. Leoni, D. Gutierrez, D. Meyers and C. Ratti  are with the SENSEable City Laboratory, Massachusetts Institute of Technology, Cambridge, MA 02139 USA.
        {\tt\small \{wweiwang, shant, leoni, davidfg, drewm, ratti\}@mit.edu}}
\thanks{W. Wang, T. Shan and D. Rus are with the Computer Science and Artificial Intelligence Lab (CSAIL),  Massachusetts Institute of Technology, Cambridge, MA 02139 USA.
    {\tt\small \{wweiwang, shant, rus\}@mit.edu}}%
}
\begin{document}
\maketitle
\thispagestyle{empty}
\pagestyle{empty}
\begin{abstract}
This paper presents a novel autonomous surface vessel (ASV), called Roboat II for urban transportation. Roboat II is capable of accurate simultaneous localization and mapping (SLAM), receding horizon tracking control and estimation, and path planning.
Roboat II is designed to maximize the internal space for transport, and can carry payloads several times of its own weight. Moreover, it is capable of holonomic motions to facilitate transporting, docking, and inter-connectivity between boats. The proposed SLAM system receives sensor data from a 3D LiDAR, an IMU, and a GPS, and utilizes a factor graph to tackle the multi-sensor fusion problem. To cope with the complex dynamics in the water, Roboat II employs an online nonlinear model predictive controller (NMPC), where we experimentally estimated the dynamical model of the vessel in order to achieve superior performance for tracking control. The states of Roboat II are simultaneously estimated using a nonlinear moving horizon estimation (NMHE) algorithm. Experiments demonstrate that Roboat II is able to successfully perform online mapping and localization, plan its path and robustly track the planned trajectory in the confined river, implying that this autonomous vessel holds the promise on potential applications in transporting humans and goods in many of the waterways nowadays.
\end{abstract}
\section{Introduction}
The increasing needs for water-based navigation in areas such as oceanic  monitoring, marine resource exploiting, and hydrology surveying have all led to strong demand from commercial, scientific, and military communities for the development of innovative autonomous vessels (ASVs)  \cite{Corke2007a, Doniec1648, Leonard1639838, paull2014auv, Dhariwal4399056, LIU201671}.
ASVs also have a promising role in the future of transportation for many coastal and riverside cities such as Amsterdam and Venice, where some of the existing infrastructures like roads and bridges are always overburdened. A fleet of eco-friendly self-driving vessels could shift the transport behaviors from the roads to waterways, possibly reducing street traffic congestion in these water-related cities.

Much progress has been made on ASV autonomy in the last several decades \cite{LIU201671}, such as localization \cite{Corke2007a,Dhariwal4399056, paull2014auv}, object detection \cite{Sukhatme5980509}, path planning \cite{Dhariwal4399056, Sukhatme21767, shan2020receding} and tracking control \cite{klinger2017control, GuerreiroSilvestre1868}. However, current ASVs are usually developed for open waters\cite{GuerreiroSilvestre1868, Manley5289429, LIU201671} and thus cannot satisfactorily meet the autonomy requirements for applications in narrow and crowded urban water environments such as Amsterdam canals.
Developing an autonomous system for vessels in urban waterways is more challenging than for traditional ASVs in open water environments.

This paper focuses on the design of localization and control problems for urban ASVs.
First, to safely navigate in urban waterways, an ASV should localize itself with centimeter-level or decimeter-level accuracy. Current ASVs usually use GPS and IMU (fused by an extended Kalman filter (EKF) or unscented Kalman filter (UKF)) which typically results in a meter-level precision \cite{LIU201671}. These GPS-IMU-based approaches can be unstable in urban waterways, where GPS signals are often severely attenuated.  A reliable multi-sensor navigation system which includes GPS, compass, speed log, and a depth sensor to account for sensor failure was proposed, but it cannot guarantee high accuracy \cite{naeem2012integrated}. To date, there is no feasible solution for accurate urban ASV localization.
Second, a number of tracking control methods such as  sliding mode method \cite{ashrafiuon2008sliding},  integrator back-stepping method \cite{khalil1996noninear, klinger2017control} and adaptive control \cite{skjetne2004nonlinear} have been proposed for ASVs. However, most of the current controllers are either verified by simulation or partly verified in open waters which do not care too much of the tracking accuracy.  Moreover, many controllers use a kinematic model instead of a dynamical one for the vessels. The control performance will always decline a lot due to the highly non-linearity of the water and the persistent disturbances in real environments.

Our recently launched Roboat project aims at developing a fleet of autonomous vessels for transportation and constructing dynamic floating infrastructure \cite{WeiICRA2018, WeiIROS2019, Mateos8793525, WangMRS2020} (e.g., bridges and stages) in the city of Amsterdam.
In our previous work \cite{WeiICRA2018, WeiIROS2019}, we have designed a quarter-scale Roboat that was able to localize itself using LiDAR and track on the reference trajectory using NMPC. By contrast, this paper develops a new large-scale vessel, Roboat II, which is capable of carrying passengers. Moreover, we develop the new SLAM algorithm, NMHE state estimation algorithm, and adapt the NMPC for large-scale Roboat II. The main contributions of our work can be summarized as follows:\\
$\tiny {\bullet}$ Designing and building of a new vessel, Roboat II;\\
$\tiny {\bullet}$ NMHE state estimation for autonomous vessels;\\
$\tiny {\bullet}$  Simultaneous localization and mapping (SLAM) algorithm for urban vessels;\\
$\tiny {\bullet}$  NMPC-NMHE control strategy with full state integration for accurate tracking;\\
$\tiny {\bullet}$  Extensive experiments to validate the developed autonomy system in rivers.

%   a complete autonomy system adapted to the Roboat architecture. Our system navigates, detects and avoids static and dynamic obstacles in a dynamic urban environment;

This paper is structured as follows. Section II overviews the Roboat prototype. Section III describes the framework of the developed autonomy system for urban ASVs. NMPC and NMHE are described in Section IV. SLAM algorithm is presented in Section V. Experiments are presented in Section VI. Section VII concludes this paper.
\section{Roboat Design}
\subsection{Hull Design}
\begin{figure}[htb]
    \centering
    \includegraphics[width=1.0\linewidth] {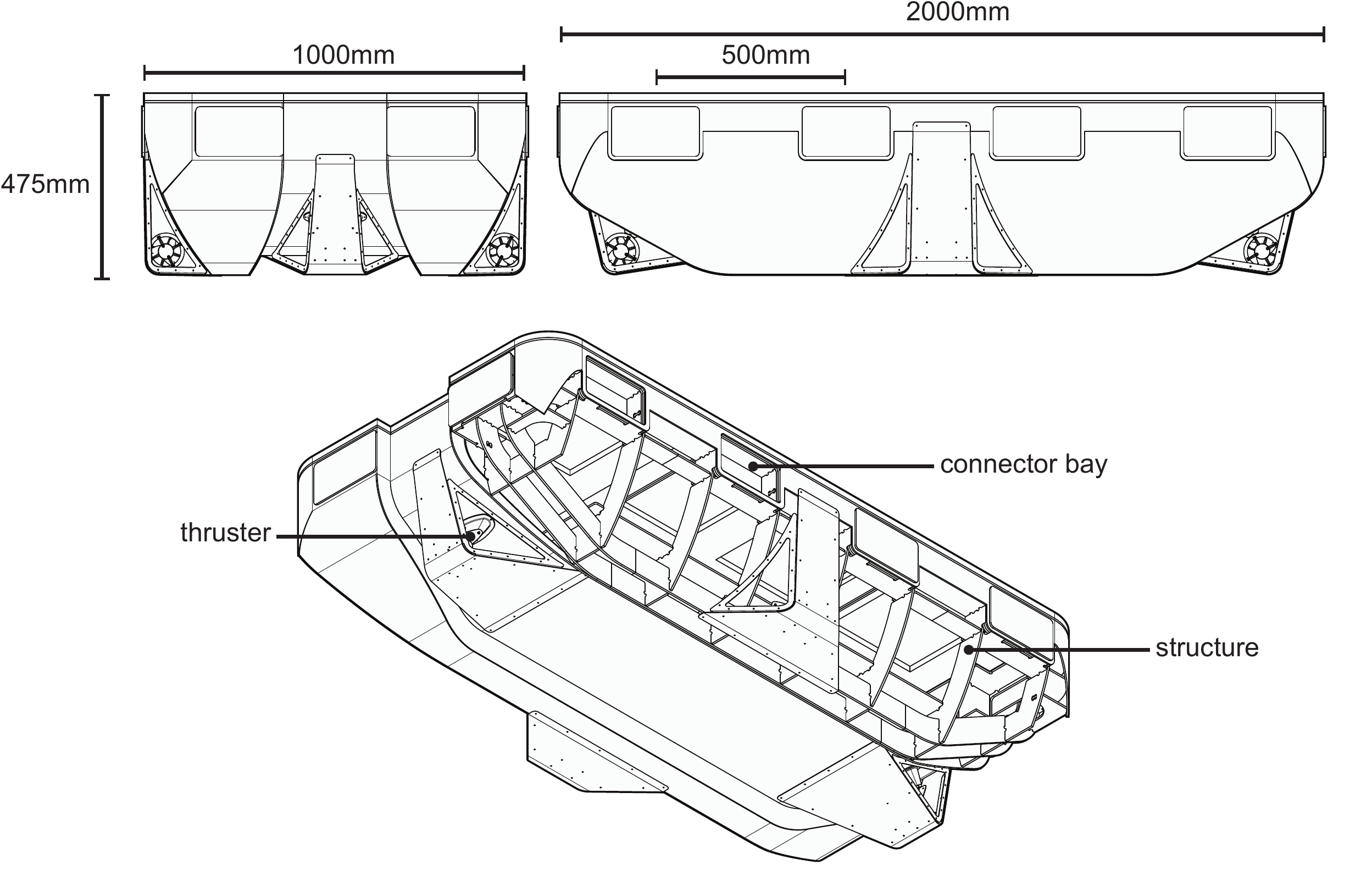}
    \caption{Mechanical design of Roboat II. }
    \label{Roboathull}
    \vspace{-3mm}
\end{figure}

Roboat II is designed starting from two main principles: simple construction and inter-connectivity between vessels. The former principle leads to define the hull shape using single-curvature surfaces.
Regarding the latter, the ability to connect multiple units in the water to build larger structures, side by side or perpendicularly, dictated the 1:2 ratio. In this way, two vessels can dock with their short sides on the long, lateral side of a third one if needed.
Also, this means an even distribution of the location of the connectors, leaving as empty bays spaced along the sides for easy accommodation of latching modules, as shown in Fig. \ref{Roboathull}.
Bolted to the structure, these modules are fast to replace, iteration after iteration, without further interventions on the hull itself.
Being those the points where the vessel is connected, they are also particularly subjected to mechanical stress. Thus, a structural rib is placed on each of its sides to distribute the forces.
With a connector every 500 mm, it is their position that dictates the internal ribs distribution.

Other than this, the need for a unit capable of maximizing the internal space and not required to move fast (cruise speed is 7.5~km/h) resulted in a bulky shape, perfectly symmetrical on the two main axes, and with the control system able to adjust the thrust to move in any desired direction.
Marine plywood has been preferred because of its lower cost and easier workability.
Similar to if we were using aluminum foil, the easier option for the assembly was to design the hull as a combination of CNC cut sheets, 4 to 6 mm thick and bent if needed along a single direction, rather than pressed to achieve double curvature.
A fiberglass coating was still applied on the exterior, with the interior space being easily accessible through 8 waterproof hatches on the top deck.
Furthermore, the whole top deck was not glued permanently to the hull but bolted instead, with a neoprene gasket in the middle.
This allows accessing the internal volume in situations where the hatches are not large enough, such as for extended intervention on the inside structure or for permanent placement of large components.
%Two variants of the hull were produced, in one exemplar each: a mono-hull and a catamaran.
%Despite this, most of the components were the same, as the catamaran version was obtained by increasing the deepness of the wide groove longitudinal to the hull on the primitive shape.
%The main difference was then the use of two couples of bow thrusters in the catamaran version, instead of only two units, to be placed in each demi-hulls instead than in a central position.
%All the thrusters were initially cover with protective grids against floating debris, but the great inefficiency measured during early tests convinced the team to get rid of these.

\begin{comment}
\begin{figure}[t]
    \centering
    \includegraphics[width=0.8\linewidth] {MasswithdraftHalfScale}
    \caption{The profile between draft and mass of the vessel. }
    \label{MasswithdraftHalfScale}
    \vspace{-3mm}
\end{figure}
\end{comment}

\subsection{Hardware}\label{hardware}
\begin{figure}[t]
    \centering
    \includegraphics[width=0.7\linewidth]{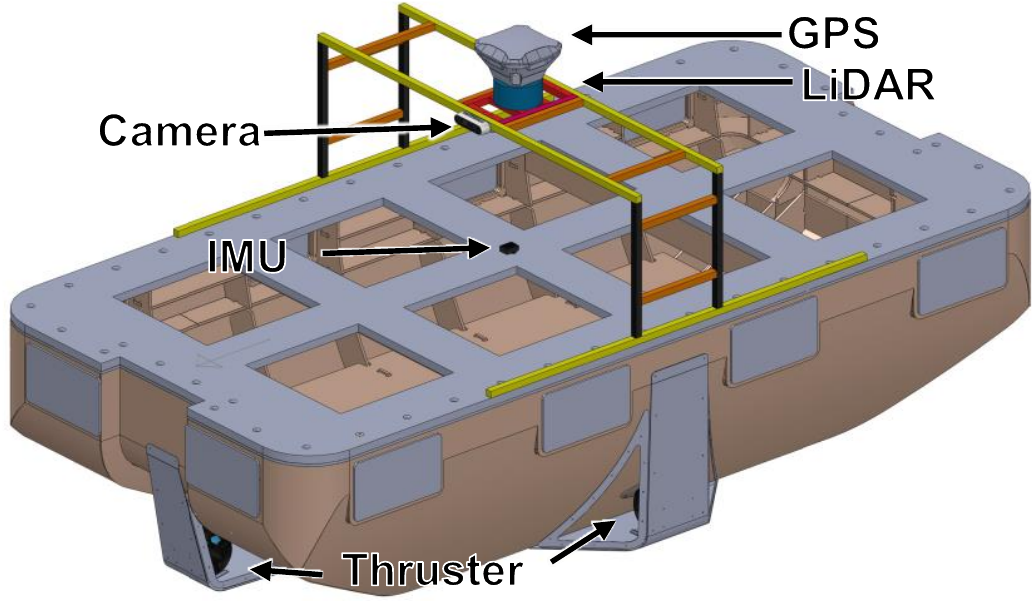}
    \caption{Hardware design of Roboat II.}
    \label{HalfscaleRoboat}
    \vspace{-3mm}
\end{figure}

We use a  small-form-factor barebone computer, Intel\textsuperscript{\textregistered} NUC (NUC7i7DNH)  as the main processor of our vessel which runs a robotics middleware, Robot Operating System (ROS). An auxiliary STM32 processor is used for converting the calculated forces from the controller to actuator signals. Roboat also has several onboard sensors, as shown in Fig. \ref{HalfscaleRoboat}. More specifications of Roboat are listed in Table \ref{technicalspecifications}.

\begin{table}[htb]
%\normalsize
\small
%\scriptsize
\begin{center}
\caption{Technical specifications of the prototype}
\label{technicalspecifications}
\begin{tabular}{ll}
\toprule[1pt]
\textbf{Items}                                      &\textbf{Characteristics}\\
\hline
Dimension  (L$\times$W$\times$H) & 2.0~m$ \times $ 1.00~m $ \times $ 0.475~m \\
Total mass                                                                         & $\sim$ 80.0~kg \\
Center of gravity height                                                                         & $\sim$ 175~mm \\
Drive mode                                                                       &Four T200 thrusters \\
Onboard sensors                                                            & 3D LIDAR, IMU, Camera, GPS \\
Power supply                                                                   & 14.8~V, 22A$\cdot$h Li-Po battery\\
Operation time                                                                & $\sim$ 2.0~hours \\
Control mode                                                                 & Autonomous/Wireless mode\\
Maximum speed                                                            & 1.0 Body Length/s\\
\hline
%\bottomrule[0.5pt]
\end{tabular}
\end{center}
\end{table}

\subsection{Hydrodynamic Analysis with CFD Simulation}
\begin{david}
\begin{comment}
%David, around half page including possible figures/tables/diagram
%important to have some beautiful simulation pictures to show our solid basis in hydrodynamics analysis
%maybe need to analyze the static and dynamic parameters of the robot (not sure if we can also calculate the added mass and drag coefficients in the simulation).
\end{comment}
The mechanical model described in the previous sections allows us to determine the hydrostatic coefficients of the prototype. Figure~\ref{fig:vol_WS} shows the mass displacement and wetted surface as functions of the vessel draught assuming it floats in brackish water (density $\sim1010$~kg/m$^3$), which have a direct impact on its dynamics.
\begin{comment}
In addition, Fig.~\ref{fig:KG_KM} displays the buoyancy and metacentric heights, $KB$ and $KM$, respectively. These values are the main stability coefficients as they characterize the righting moment ($\mathcal{M}$) for the expected small roll and pitch angles, such that
\begin{align}
\mathcal{M} &= \left(KM - KG\right) \,m \sin{\theta} \,,
\end{align}{}
%
where $KG$ is the center of gravity height that depends on the loading condition, $m=\rho \mathcal{V}$ is the mass displaced, $\rho$ the fluid density, and $\theta$ the roll/pitch angle.
\end{comment}
%
\begin{figure}[htb]
    \centering
    \includegraphics[width=0.7\linewidth]{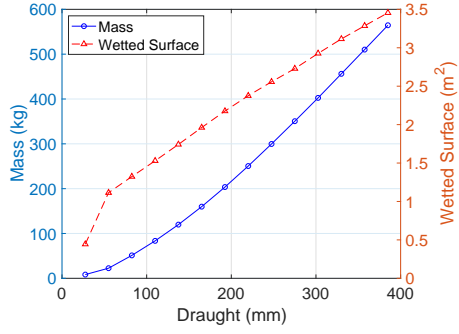}
    \caption{Displaced mass and wetted surface values as function of draught.}
    \label{fig:vol_WS}
    \vspace{-3mm}
\end{figure}
%\begin{figure}[htb]
%    \centering
%    \includegraphics[width=0.9\linewidth]{Figures/hydrostatics_KB_KM.eps}
%    \caption{Center of buoyancy and metacentric heights as function of the prototype draught}
%    \label{fig:KG_KM}
%\end{figure}

Based on the total mass and center of gravity height of the tested prototype from Table~\ref{technicalspecifications}, the expected draught of the prototype from Fig.~\ref{fig:vol_WS} is 107~mm.
%We can also determine the increase in mass displacement per millimeter increment of draught, $\Delta m_{1\text{mm}} \sim 1.25$~kg, and the righting moment per degree of roll and pitch, $\mathcal{M}_{1^{\circ},\text{roll}} \sim 1.1$~kg$\cdot$m and $\mathcal{M}_{1^{\circ},\text{pitch}} \sim 4.4$~kg$\cdot$m, respectively.
Given that these Roboat units will operate in environments with wave heights below 10~cm, which yield roll and pitch moments $\sim$5 and $\sim$10 kg$\cdot$m, respectively, we expect roll and pitch angles below 5$^{\circ}$. These outcomes validate the assumption of planar motion followed in the dynamic model described in Sec.~\ref{ss:dynamic_model}.
Moreover, \ref{fig:vol_WS}  also shows the capability to carry a much larger payload than the quarter-scale model \cite{WeiICRA2018}. Roboat II can carry two people on-board (see supplementary material).
%
%Table~\ref{tab:hydrodynamics} summarizes the stability coefficients for this floating condition, where $\Delta m_{1\text{mm}}$ indicates the increase in mass displaced per millimeter increment of draught, and $\mathcal{M}_{1^{\circ}}$ the righting moment per degree of roll/pitch. %The values obtained allow us to expect small disturbance loads induced by the predicted waves it will encounter when working in urban environments.
%\begin{table}[htb]
%\normalsize
%\small
%\scriptsize
%\begin{center}
%\caption{Hydrostatic characteristics of the prototype for simulation}
%\label{tab:hydrodynamics}
%\begin{tabular}{ll}
%\toprule[1pt]
%\textbf{Items}                                      &\textbf{Characteristics}\\
%\hline
%Dimension (L$\times$W$\times$H) & 2.0~m$ \times $ 1.00~m $ \times $ 0.45~m \\
%$m$ & 80.0~kg \\
%$KG$ & 175~mm \\
%KM_{\text{roll}}$ / $KM_{\text{pitch}}$ & 0.95~m / 3.15~m \\
%$\Delta m_{1\text{mm}}$ & 1.25~kg \\
%$\mathcal{M}_{1^{\circ},\text{roll}}$ / $\mathcal{M}_{1^{\circ},\text{pitch}}$ & 1.1~kg$\cdot$m / 4.4~kg$\cdot$m \\
%\hline
%%\bottomrule[0.5pt]
%\end{tabular}
%\end{center}
%\end{table}

Additionally, we used the SolidWorks\textsuperscript{\textregistered} Flow Simulation package to evaluate numerically the hydrodynamic response of the Roboat under the design operating conditions. Figure~\ref{fig:CFD_sim} shows a side view with the pressure distribution over the hull as the Roboat moves forward at 0.8~m/s, as well as the velocity streamlines around it. The results show how the smooth hull efficiently deflects the incident flow, minimally perturbing the free surface and avoiding high dynamic pressure concentrations. %Furthermore, integrating the computed forces over the hull, we predict a drag force at this condition of ... N, which can be transferred to the linearized dynamical model used in our control scheme described in Sec... as a damping coefficient of ...
\begin{figure}[htb]
    \centering
    \includegraphics[width=1.0\linewidth]{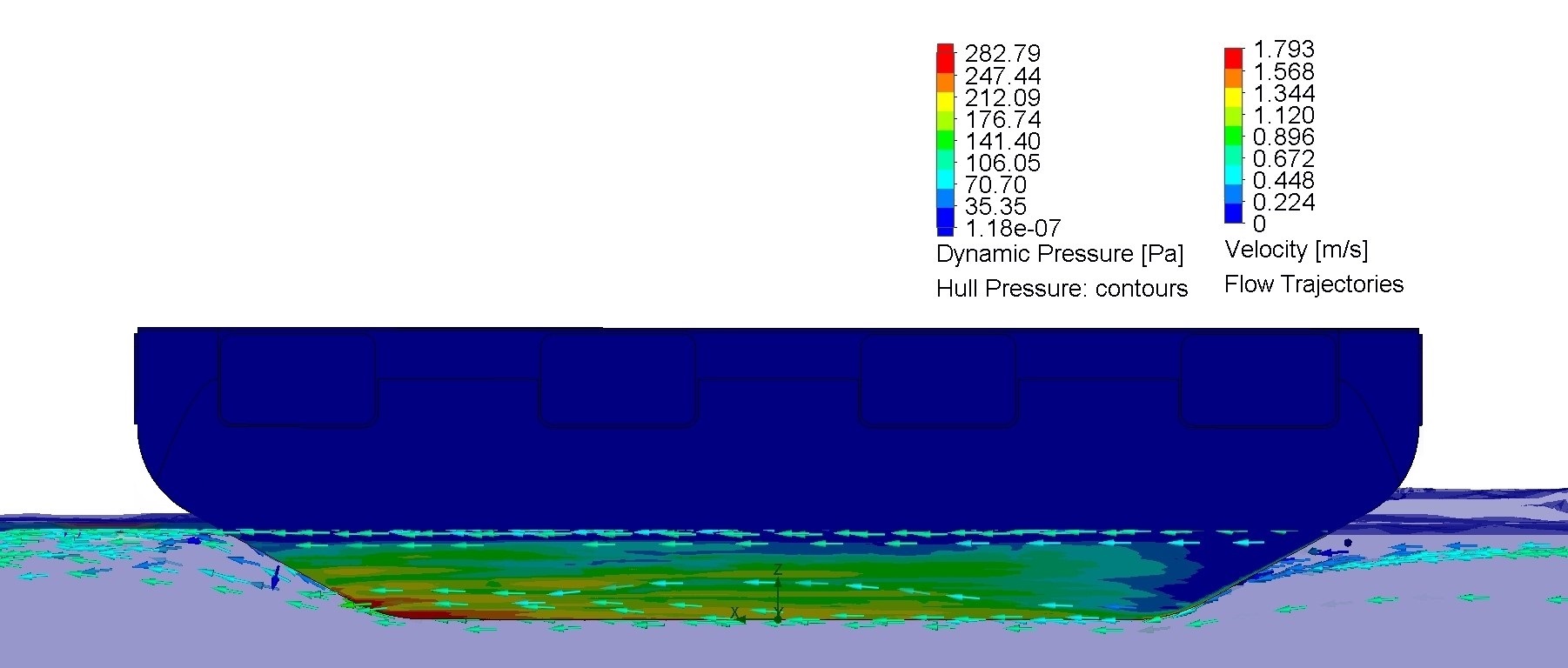}
    \caption{CFD results from SolidWorks\textsuperscript{\textregistered} Flow Simulation with the Roboat moving forward at 0.8~m/s.}
    \label{fig:CFD_sim}
    \vspace{-3mm}
\end{figure}
\end{david}

\section{Autonomy Framework of Urban Vessels}
\begin{figure}[htb]
    \centering
    \includegraphics[width=0.9\linewidth] {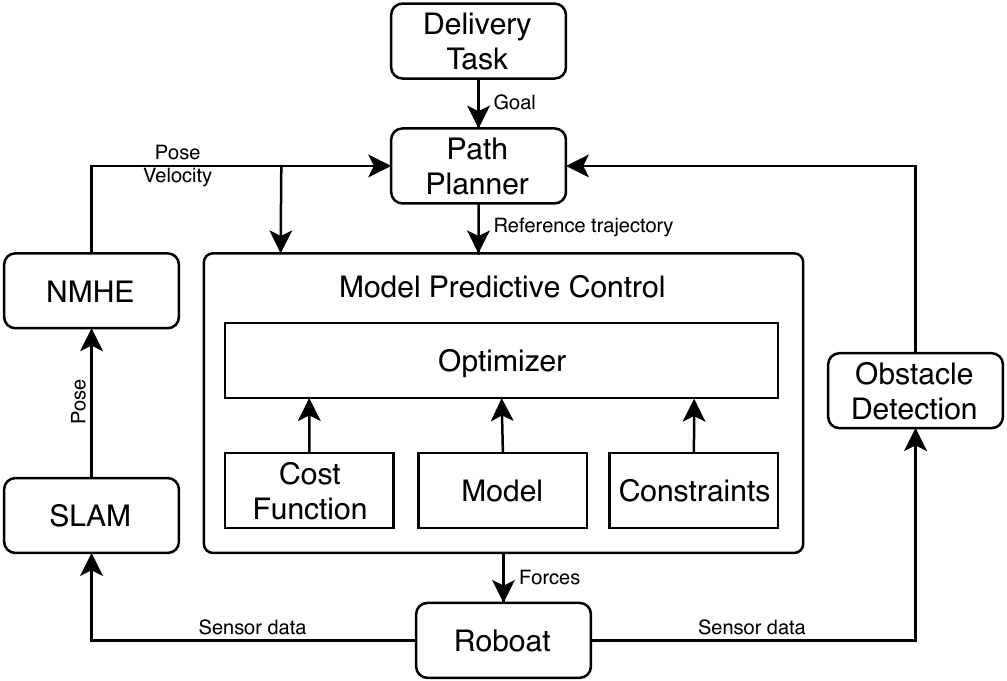}
    \caption{Current autonomy framework of Roboat II. It mainly contains a planner,a SLAM module, an NMPC tracking module, an NMHE state estimator, and a simple object detector. }
    \label{AutonomyFramework}
    \vspace{-3mm}
\end{figure}

In this section, we describe the autonomy framework of our Roboat II which can move in urban waterways, as is shown in Fig. \ref{AutonomyFramework}.
The autonomous Roboat conducts a transportation task in the canals as follows.  When a task such as passenger delivery is required from a user at a specific position, the system coordinator will assign this delivery task to an unoccupied Roboat that is closest to the passenger. When Roboat picks up the passenger, it will first plan its path to the destination and then generate a feasible path based on the current traffic condition. Then, Roboat starts to localize itself by running the developed SLAM algorithm which utilizes LiDAR, IMU, and GPS sensors. The NMPC controller will accurately and robustly track the reference trajectories from the planner during the whole task in an urban waterway. The planner will avoid obstacles if it receives obstacle information from an obstacle detector. We use a simple A$^{\star}$ planner for the path generation and focus on SLAM, dynamics modeling and identification, NMPC tracking and NMHE state estimation in this study.
\section{Receding  Horizon  Control and Estimation}
Nonlinear model predictive control (NMPC) is a dynamic optimization-based strategy for feedback control which determines the current control action by optimizing the system behavior over a finite window, often referred to as the prediction horizon.
The NMPC controller is responsible for tracking the calculated optimal trajectory from the path planner. The states of the vessel are simultaneously estimated using a nonlinear moving horizon estimation (NMHE) algorithm.
These estimated states are further fed to the NMPC, as shown in Fig. \ref{AutonomyFramework}. The performance of the NMPC largely relies on the selected system model. Therefore, we build a dynamical model for Roboat II, and experimentally identify the unknown parameters in the model to achieve superior performance for NMPC control.
\subsection{Dynamical Model}\label{ss:dynamic_model}
The dynamics of our vessel is described by the following nonlinear differential equation \cite{WeiICRA2018}
\begin{eqnarray}\label{MFCPredynamics}
&&\dot{\mathbf{x}}=\mathbf{T}(\mathbf{x})\mathbf{v}  \label{MFCPredynamicsA}\\
&&\dot{\mathbf{v}}=\mathbf{M}^{-1}(\mathbf{\tau}+\mathbf{\tau}_{\text{env}})-\mathbf{M}^{-1}(\mathbf{C}(\mathbf{v})+\mathbf{D}(\mathbf{v}))\mathbf{v} \label{MFCPredynamicsB}
\end{eqnarray}
where  $\mathbf{x}=[x~y~\psi]^{T} \in\mathbb{R}^{3\times1}$ is the position and heading angle of the vessel in the inertial frame; $\mathbf{v}=[u~v~r]^{T}\in \mathbb{R}^{3\times1}$ denotes the vessel velocity, which contains the surge velocity ($u$), sway velocity ($v$), and yaw rate ($r$) in the body fixed frame; $\mathbf{T}(\mathbf{x})\in \mathbb{R}^{3\times3}$ is the transformation matrix converting a state vector from body frame to inertial frame; $\mathbf{M} \in \mathbb{R}^{3\times3}$ is the positive-definite symmetric added mass and inertia matrix; $\mathbf{C}(\mathbf{v})\in\mathbb{R}^{3\times3}$ is the skew-symmetric vehicle matrix of Coriolis and centripetal terms; $\mathbf{\tau}_{\text{env}} \in\mathbb{R}^{3\times1}$ is the environmental disturbances from the wind, currents and waves; $\mathbf{D}(\mathbf{v})\in\mathbb{R}^{3\times3}$ is the positive-semi-definite drag matrix-valued function; $\mathbf{\tau}=[\tau_u~ \tau_v~ \tau_r]^{T} \in\mathbb{R}^{3\times1}$ is the force and torque applied to the vessel in all three DOFs, which is defined as follow
\begin{eqnarray}\label{AppliedForceMaxtrix}
\mathbf{\tau}
=\mathbf{B}\mathbf{u}
=
\left[
 \begin{array}{cccc}
1                      &  1                                      &    0                       & 0\\
0                      &  0                                      &    1                       & 1\\
\dfrac{a}{2}&-\dfrac{a}{2}                 &     \dfrac{b}{2} &-\dfrac{b}{2}
\end{array}
\right]
 \left(
 \begin{array}{c}
f_1\\
f_2\\
f_3\\
f_4
\end{array}
\right)
\end{eqnarray}
where $\mathbf{B}\in\mathbb{R}^{4\times3}$ is the control matrix describing the thruster configuration and $\mathbf{u}=[f_1~f_2~f_3~f_4]^{T}\in\mathbb{R}^{4\times1}$ is the control vector where $f_1$, $f_2$, $f_3$ and $f_4$ represent the left, right, anterior, and rear thrusters, respectively; $a$ is the distance between the transverse propellers and $b$ is the distance between the longitudinal propellers.
%Fig. \ref{RobotFrame_HB} illustrates the two coordinate systems and the thruster forces acting on the vehicle.
$\mathbf{M}$, $\mathbf{C}(\mathbf{v})$ and  $\mathbf{D}(\mathbf{v})$ are mathematically described as follows:
\begin{eqnarray}
\mathbf{M}=\text{diag}\{m_{11}, m_{22}, m_{33}\} \label{InertiaMaxtrix}
\end{eqnarray}
\begin{eqnarray}
\mathbf{C}(\mathbf{v})=
 \left[
 \begin{array}{ccc}
0                                   &  0                               & -m_{22}v\\
0                                   &0                                 & m_{11}u\\
m_{22}v               &-m_{11}u               &0
\end{array}
\right]
\end{eqnarray}
\begin{eqnarray}
 \mathbf{D}(\mathbf{v})=\text{diag} \{X_u, Y_v, N_r\}\label{LinearDrag}
\end{eqnarray}

Further, by combining (\ref{MFCPredynamicsA}) and (\ref{MFCPredynamicsB}), the complete dynamic model of the vessel is reformulated as follow
\begin{eqnarray}\label{MPCdynamics}
\dot{\mathbf{q}}(t)=f(\mathbf{q}(t),\mathbf{u}(t))
\end{eqnarray}
where $\mathbf{q}=[x~y ~\psi~ u~ v~ r]^{T}\in \mathbb{R}^{6\times1}$  is the state vector of the vessel, and $f(\cdot, \cdot, \cdot): \mathbb{R}^{n_q}\times \mathbb{R}^{n_u} \times \mathbb{R}^{n_p}\longrightarrow \mathbb{R}^{n_q}$ denote the continuously differentiable state update function. The system model describes how the full state $\mathbf{q}$  changes in response to applied control input $\mathbf{u} \in\mathbb{R}^{4\times1}$. Similarly, a nonlinear measurement model denoted $\mathbf{h}(t)$ can be described with the following equation:
\begin{eqnarray}\label{measurementequation}
\mathbf{z}(t)=h(\mathbf{q}(t),\mathbf{u}(t))
\end{eqnarray}
where $\mathbf{z}=[ x~y ~\psi~ r~f_1~f_2~f_3~f_4] \in \mathbb{R}^{8\times1}$ denotes the measurement vector,  and $h(\cdot, \cdot):  \mathbb{R}^{n_q}\times  \mathbb{R}^{n_u} \longrightarrow \mathbb{R}^{n_z}$ denote measurement function.

Next, the unknown hydrodynamic parameter vector, $\xi=[m_{11}~m_{22}~m_{33}~X_u~Y_v~N_r]^{T}$, in the dynamical model is required to be identified before applying it to the controller.
The estimation of $\xi$ using the experimental data set ${\mathbf{v}^{\text{s}},\mathbf{u}^{\text{s}}}$ is a grey-box identification problem. The identification can be treated as an optimization problem described below
\begin{subequations}\label{modelidentificationformulation}
\begin{eqnarray}
&&\!\!\!\!\!\!\!\!\!\!\!\!\!\!\!\!\!\!\!\!\!\!\!\!\!\!\!\!\!\!\!\!\!\underset{\xi}{\text{min}}~{\sum}_{t=0}^{T_s} \varepsilon(t)^{T}w\varepsilon(t),\\
\text{s.t.}~~ &&\xi_{l} \leq \xi \leq \xi_{u}, \\
&&\mathbf{q}(t)=f(\mathbf{q}(t),\mathbf{u}(t),\xi),  t\in[0 ~T_s],
\end{eqnarray}
\end{subequations}
where $\varepsilon(t)\in\mathbb{R}^{3\times1}$ is the deviation between the experimental velocity $\mathbf{v}^s(t)$ and the simulated velocity $\mathbf{v}^m(t)$ at time $t$.
$\xi_l$ and $\xi_u$ are the lower and upper bounds of $\xi$, respectively. $w\in\mathbb{R}^{3\times3}$ represents the weight matrix for the optimization. Different from our previous work \cite{WeiICRA2018}, we adopt Sequential Quadratic Programming (SQP) method to numerically solve (\ref{modelidentificationformulation}) in this study because SQP satisfies bounds at all iterations and  (\ref{modelidentificationformulation}) is not a large-scale optimization problem.
%Moreover, previous studies show that SQP outperforms the other methods in terms of efficiency, accuracy, and percentage of successful solutions, over a large number of nonlinear problems \cite{schittkowski1986nlpql}.
%Specifically, SQP allows us to apply Newton's method for constrained optimization. At each major iteration, an approximation is made of the Hessian of the Lagrangian function using a quasi-Newton updating method. This is then used to generate a QP sub-problem whose solution is used to form a search direction for a line search procedure.

%\begin{table}[ht]
%\small
%\begin{center}
%\caption{Results of Hydrodynamic Parameter Identification}
%\label{IdentifiedParameter}
%\begin{tabular}{ccccccc}
%\toprule[1.2pt]
%\textbf{Item}              &$m_{11}$       & $m_{22}$       &$m_{33}$      &$X_{u}$      &$Y_{v}$                 &$N_{r}$   \\
%\toprule[0.8pt]
%                        Value     &12.982         & 23.318          &1.273          &6.012        &7.112                  &0.771 \\
%\bottomrule[1pt]
%\end{tabular}
%\end{center}
%\end{table}
\subsection{Nonlinear Model Predictive Control} \label{NMPCsection}
For our  trajectory tracking problem involving online dynamics learning, we formulate the optimal control problem for NMPC in the form of a least square function to penalize the deviations of predicted state ($\mathbf{q}_k$) and control ($\mathbf{u}_k$) trajectories from their specified references, over the given prediction horizon window $N_c$ ($t_j\leq t \leq t_{j+N_c}$):
\begin{subequations} \label{OCP}
\begin{eqnarray}
\underset{\mathbf{q}_k,\mathbf{u}_k}{\text{min}}\dfrac{1}{2}\Big\{\sum_{k=j}^{j+N_c-1}(\|\mathbf{q}_k-\mathbf{q}_k^{\text{ref}}\|^2_{W_q}+\|\mathbf{u}_k-\mathbf{u}_k^{\text{ref}}\|^2_{W_u})+ \label{OCPequation}\\
\|\mathbf{q}_{N_c}-\mathbf{q}_{N_c}^{\text{ref}}\|^2_{W_{N_c}}\Big\} \nonumber\\
\text{s.t.}~~\mathbf{q}_j= \hat{\mathbf{q}}_j,\\
\mathbf{q}_{k+1}=f(\mathbf{q}_k,\mathbf{u}_k),  k=j,\cdot \cdot \cdot, j+N_c-1,\\
\mathbf{q}_{k,\text{min}}\leq\mathbf{q}_k\leq \mathbf{q}_{k,\text{max}}, k=j,\cdot \cdot \cdot, j+N_c,\\
\mathbf{u}_{k,\text{min}}\leq\mathbf{u}_k\leq \mathbf{u}_{k,\text{max}}, k=j,\cdot \cdot \cdot, j+N_c-1,
\end{eqnarray}
\end{subequations}
 where $\mathbf{q}_k \in \mathbb{R}^{n_q}$ denotes the vessel state,  $\mathbf{u}_k \in \mathbb{R}^{n_u}$ denotes the control input, $ \hat{\mathbf{q}}_j \in \mathbb{R}^{n_q}$ denotes the current state estimate, $\mathbf{q}_k^{\text{ref}}$ and $\mathbf{u}_k^{\text{ref}}$ denote the time-varying state and control references, respectively; $\mathbf{q}_{N_c}^{\text{ref}}$ denotes the terminal state reference; $W_q\in \mathbb{R}^{n_q\times n_q}$, $W_u\in \mathbb{R}^{n_u\times n_u}$, and $W_{N_c}\in \mathbb{R}^{n_q\times n_q}$ are the positive definite weight matrices that penalize deviations from the desired values. These weight matrices are assumed constant for a certain scale vessel in this study. Moreover, $\mathbf{p}$ is the parameter vector in the model which is referred as the payload of the vessel in this study.  Furthermore,  $\mathbf{q}_{k,\text{min}}$ and $\mathbf{q}_{k,\text{max}}$ denote the lower and upper bounds of the states, respectively; $\mathbf{u}_{k,\text{min}}$ and $\mathbf{u}_{k,\text{max}}$ denote the lower and upper bounds of the control input, respectively.

The weighting matrices $W_q$, $W_u$ and $W_{N_c}$ for the NMPC used in the experiments are selected as
\begin{eqnarray}
 &&W_q=\text{diag}\{200,200,100,10,10,10\}\\
&&W_{N_c}=\text{diag}\{1000,1000,500,50,50,150\} \\
&& W_u=\text{diag}\{1,1,1,1\}
 \end{eqnarray}
The prediction horizon $N_c = 4$ s,  and the constraints on the control input $\mathbf{u}$ used in the experiments are chosen as follow
\begin{eqnarray}\label{forcelimits}
\mathbf{-50}_{4\times 1}~\text{N} \leq\mathbf{u}_{\text{quart}}\leq \mathbf{50}_{4\times 1} ~\text{N}
\end{eqnarray}
\subsection{Nonlinear Moving Horizon Estimation}
Online state estimation is employed to further generate more accurate states considering the dynamics and state constraints of the vessel.
Nonlinear Moving Horizon Estimation (NMHE) is an online optimization-based state estimation approach that can handle nonlinear systems and satisfy inequality constraints on the estimated states and parameters \cite{Rao1178905}.
In a similar manner, we utilize a least square function to penalize the deviation of estimated outputs from the measurements, formulating the NMHE problem as follows: At current time $t_j$ there shall be $N_c$ measurements $\mathbf{z}_{j-M+1}, ..., \mathbf{z}_j \in \mathbb{R}^{n_z}$ available, associated to the time instants $t_{j-N_c+1}< ...< t_j$ in the past. $T_{\text{E}}=t_j-t_{j-N_c+1}$ is the length of the  horizon. Finally, the discrete time dynamic optimization problem to estimate the constrained states ($\hat{\mathbf{q}}$) at time $t_j$ using the available measurements within the horizon, is solved by
\begin{subequations} \label{ObjectiveFunctionMHE}
\begin{eqnarray}
\underset{\hat{\mathbf{q}}_k}{\text{min}}\|\hat{\mathbf{q}}_{j-N_c+1}-\bar{\mathbf{q}}_{j-N_c+1})\|^{2}_{P_L}\!\!+\!\!\!\!\!\!\sum^{j}_{k=j-N_c+1}\!\!\!\!\|\mathbf{z}_k-\mathbf{h}(\hat{\mathbf{q}}_k,\mathbf{u}_k)\|^{2}_{R_T}\\
\text{s.t.}~
\hat{\mathbf{q}}_{k+1}=f(\hat{\mathbf{q}}_{k+1},\mathbf{u}_k)+\mathbf{w}_k,  k=j-N_c+1,..., j-1, \\
\hat{\mathbf{q}}_{k,\text{min}}\leq\hat{\mathbf{q}}_k\leq \hat{\mathbf{q}}_{k,\text{max}}, k=j-N_c+1,..., j,
\end{eqnarray}
\end{subequations}
where $\hat{\mathbf{q}}_{k,\text{min}}$ and $\hat{\mathbf{q}}_{k,\text{max}}$ denote the lower and upper bounds on the estimated states of the vessel, respectively. The weighting matrices $P_{N_c}$ and $R_k$ are usually interpreted as the inverses of the measurement and process noise covariance matrices, respectively. $P_{N_c}$ and $R_k$ should be selected adequately to achieve good state estimates based on the knowledge or prediction of the error distributions. Moreover, $\bar{\mathbf{q}}_{j-N_c+1}$ represents the estimated state at the start of estimation horizon $t_{j-N_c+1}$.

The weighting matrices $P_L$ and $R_T$ are the inverses of the process and measurement noise covariance matrices, respectively. Considering the noise characteristics of sensors listed in Section \ref{hardware}, the weighting matrix $R_T$ used in the experiments is chosen as follow
\begin{eqnarray}
\!\! \!\!R_T\!\!\!\!\!\!&\!\!=\!\!&\!\!\!\!\!\!\text{diag}\{\sigma_{x}^{2},\sigma_{y}^{2},\sigma_{\psi}^{2},\sigma_{r}^{2},\sigma_{f_1}^{2},\sigma_{f_2}^{2},\sigma_{f_3}^{2},\sigma_{f_4}^{2}\}^{-1}  \nonumber \\
 \!\!\!\!\!\!&=&\!\!\!\!\!\!\text{diag}\{0.0005,0.0005,0.0005,0.0001,1, 1, 1, 1\}^{-1}
 \end{eqnarray}
Moreover, the weighting matrix $P_L$ used in the experiments are selected as follow
 \begin{eqnarray}
\!\! \!\!P_L\!\!\!\!\!\!&\!\!=\!\!&\!\!\!\!\!\!\text{diag}\{\sigma_{x}^{2},\sigma_{y}^{2},\sigma_{\psi}^{2},\sigma_{u}^{2},\sigma_{v}^{2},\sigma_{r}^{2}\}^{-1}  \nonumber \\
 \!\!\!\!\!\!&=&\!\!\!\!\!\!\text{diag}\{1, 1, 1, 0.1, 0.1, 1\}^{-1}
 \end{eqnarray}
\section{Simultaneous Localization and Mapping}
\label{sec::localization}

% define some symbols
\newcommand{\bomega}{\boldsymbol{\omega}}
\newcommand{\bx}{\bold{x}}
\newcommand{\bq}{\bm{q}}
\newcommand{\bb}{\bold{b}}
\newcommand{\bn}{\bold{n}}
\newcommand{\ba}{\bold{a}}
\newcommand{\bg}{\bold{g}}
\newcommand{\bv}{\bold{v}}
\newcommand{\bp}{\bold{p}}
\newcommand{\bR}{\bold{R}}
\newcommand{\bTR}{\bold{{}^{\mathsf{T}}R}}
\newcommand{\bBW}{\bold{BW}}
\newcommand{\Dt}{\Delta t}
\newcommand{\bT}{\bold{T}}
\newcommand{\bTT}{\bold{{}^{\mathsf{T}}T}}
\newcommand{\bd}{\bold{d}}
\newcommand{\bM}{\bold{M}}
\newcommand{\bpoint}{\bold{p}}
\newcommand{\textF}{\text{F}}
\newcommand{\primeF}{'\textF}
\newcommand{\bigbar}{\Big|}
\newcommand{\bfracNoLine}[2]{\genfrac{|}{|}{0pt}{0}{#1}{#2}}

\begin{figure*}[ht!]
	\centering
	\includegraphics[width=.95\textwidth]{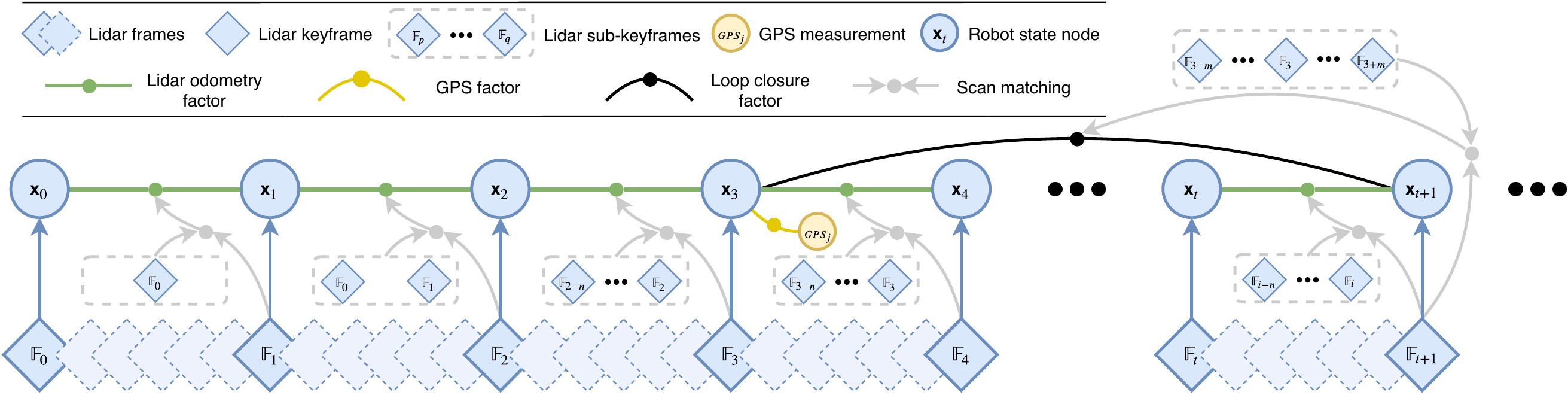}
	\caption{The framework of the SLAM system. The system receives input from a 3D lidar, an IMU, and a GPS. Three types of factors are introduced to construct the factor graph.: (a) lidar odometry factor, (c) GPS factor, and (c) loop closure factor. The generation of these factors is discussed in Section \ref{sec::localization}.}
	\label{fig::factor-graph}
	\vspace{-3mm}
\end{figure*}

An overview of the proposed SLAM system, which is adapted from \cite{shan2020lio}, is shown in Fig. \ref{fig::factor-graph}. The system receives sensor data from a 3D lidar, an IMU, and a GPS. SLAM algorithm aims to estimate the state $\mathbf{x}=[x~y~\psi]\in\mathbb{R}^{3\times1}$ of the vessel given the sensor measurements. This state estimation problem can be formulated as a posteriori (MAP) problem. To seamlessly incorporate measurements from various sensors, we utilize a factor graph to model this problem. Then solving the MAP problem is equivalent to solving a nonlinear least-squares problem \cite{dellaert2017factor}.

We introduce three types of $factors$ along with one $variable$ type for factor graph construction. This variable, representing the vessel's state, is referred to as $nodes$ of the graph. The three types of factors are: (a) lidar odometry factors, (b) GPS factors, and (c) loop closure factors. To limit the memory usage and improve the efficiency of the localization system, we add new node $\bx \in\mathbb{R}^{3\times1}$ to the graph using a simple but effective heuristic approach. A new node $\bx$ is only added when the position or rotation change of the vessel exceeds a user-defined threshold. We use incremental smoothing and mapping with the Bayes tree (iSAM2) \cite{kaess2012isam2} to optimize the factor graph upon the insertion of a new node. The process for generating the aforementioned factors is described in the following sections.

% \begin{figure}[ht]
% 	\centering
% 	\subfigure[Camera footage]{\includegraphics[width=.49\columnwidth]{Figures/demo-camera.png}}
% 	\subfigure[Velodyne point cloud]{\includegraphics[width=.49\columnwidth]{Figures/demo-velodyne.png}}
% 	\subfigure[Extracted features]{\includegraphics[width=.49\columnwidth]{Figures/demo-feature.png}}
% 	\subfigure[Point cloud map]{\includegraphics[width=.49\columnwidth]{Figures/demo-map.png}}
% 	\caption{Visualization of the feature extraction and the point cloud map. (a) A camera footage that shows the operating environment. (b) A point cloud data from the 3D lidar. (c) Extracted edge and planar features from (b) are colored as green and pink points respectively. (d) The resulting point cloud map of the canal, where each white square indicates a node in the factor graph. In (b) and (c), changes in color indicate elevation change.}
% 	\label{fig::point-cloud-demo}
% \end{figure}

%%%%%%%%%%%%%%%%%%%%%%%%%%%%%
\subsection{Lidar Odometry Factor}
\label{sec::lidar-odometry}

We perform feature extraction using the raw point cloud from the 3D lidar. Similar to the process introduced in \cite{LEGO-LOAM}, we first project the raw point cloud onto a range image as the arriving point cloud may not be organized. Each point in the point cloud is associated with a pixel in the range image. Then we calculate the roughness value of a pixel in the range image using its neighboring range values.
% Let $S$ be the continuous pixels of a pixel $p_{i}$ in the same row of the image. The same amount of neighboring pixels are selected on the left and right side $p_{i}$. We calculate the roughness $c_i$ of pixel $p_{i}$ in $S$,
% \begin{align}
% c_i = \frac{1}{|S| \cdot \| r_{i}\|} \mathlarger{\mathlarger{\|}} \underset{\mathsmaller{j \in S, j \neq i}}{\mathlarger{\mathlarger{\Sigma}}}{(r_{j}-r_{i})} \mathlarger{\mathlarger{\|}},
% \label{eq::roughness}
% \end{align}
% where $r_{i}$ indicates the range value of pixel $p_{i}$. The size of $S$, $|S|$, is chosen to be 10 here. Edge and planar features are extracted by evaluating the roughness value from Equation \ref{eq::roughness}.
Points with a large roughness value are classified as edge features. Similarly, a planar feature is determined by a small roughness value. We denote the extracted edge and planar features from a lidar scan at time $t$ as $\textF^{e}_{t}$ and $\textF^{p}_{t}$ respectively. All the features extracted at time $t$ compose a lidar \textit{frame} $\mathbb{F}_{t}$, where $\mathbb{F}_{t} = \{\textF^{e}_{t}, \textF^{p}_{t}\}$. Note that $\mathbb{F}_{t}$ is represented in the local sensor frame. A more detailed description of the feature extraction process can be found in \cite{LEGO-LOAM}.

It is computationally intractable to add factors for every lidar frame. Thus we adopt the concept of \textit{keyframe} selection, which is widely used in the visual SLAM field. We select a lidar frame $\mathbb{F}_{t+1}$ as a keyframe when the vessel's pose change exceeds a user-defined threshold. In this paper, we select a new lidar keyframe when the vessel's position change exceeds $1m$ or the rotation change exceeds $10^\circ$. The lidar frames between two keyframes are discarded. When a lidar keyframe is selected, we perform scan-matching to calculate the relative transformation between the new keyframe with the previous sub-keyframes. The sub-keyframes are obtained by transforming the previous $n$ keyframes into the global world frame. These transformed keyframes are then merged together into a voxel map $\bM_{t}$. Note that $\bM_{t}$ is composed of two sub-voxel maps because we extract two types of features in the previous step. These two sub-voxel maps are denoted as $\bM^{e}_{t}$, the edge feature voxel map, and $\bM^{p}_{t}$, the planar feature voxel map. The relationships between the feature sets and the voxel maps can be represented as:
\begin{align}
	\nonumber
	&\bM_{t} = \{\bM^{e}_{t}, \bM^{p}_{t}\} \\
	\nonumber
	&where: \bM^{e}_{t} =\; \primeF^{e}_{t} \cup \primeF^{e}_{t-1} \cup...\cup\primeF^{e}_{t-n} \\
	\nonumber
	&\;\;\;\;\;\;\;\;\;\;\;\;\;\; \bM^{p}_{t} =\; \primeF^{p}_{t} \cup \primeF^{p}_{t-1} \cup ... \cup \primeF^{p}_{t-n},
\end{align}
where $\primeF^{e}_{t}$ and $\primeF^{p}_{t}$ are the transformed edge and planar features in the global frame. $n$ is chosen to be 25 for all the tests. The new lidar keyframe $\mathbb{F}_{t+1}$ is transformed into the global world frame using the initial guess from IMU. The transformed new keyframe in the world frame is denoted as $'\mathbb{F}_{t+1}$. With $'\mathbb{F}_{t+1}$ and the voxel map $\bM_{t}$, we perform scan-matching using the method proposed in \cite{LOAM} and obtain the relative transformation $\Delta\bT_{t,\;t+1}$ between them. At last, $\Delta\bT_{t,\;t+1}$ is added as the lidar odometry factor into the factor graph.

%%%%%%%%%%%%%%%%%%%%%%%%%%%%%%%%%
\subsection{GPS Factor}
\label{sec::gps}

Though we can achieve low-drift state estimation solely by using lidar odometry factors, the localization system still suffers from drift during long-duration navigation tasks. We thus utilize the absolute measurements from GPS and incorporate them as GPS factors into the factor graph. When the GPS measurements are available, we first transform them into the Cartesian coordinate frame using the method proposed in \cite{moore2016generalized}. Upon the insertion of a new node to the factor graph, we then add a new GPS factor and associate it with the new node.

%%%%%%%%%%%%%%%%%%%%%%%%%%%%
\subsection{Loop Closure Factor}
\label{sec::loop-closure}

Due to the utilization of a factor graph, loop closures can also be seamlessly incorporated into the proposed system. Successful detection of loop closure is introduced as a loop closure factor in the factor graph. For illustration, we introduce a naive but effective loop closure detection approach that is based on Euclidean distance. When a new state $\bx_{i+1}$ is added into the factor graph, as is shown in Fig. \ref{fig::factor-graph}, we search the graph and find the prior states that are within a certain distance to $\bx_{i+1}$. For example, $\bx_3$ is the returned candidate state. We then extract sub-keyframes $\{\mathbb{F}_{3-m}, ..., \mathbb{F}_{3}, ..., \mathbb{F}_{3+m}\}$ and merge them into a local voxel map, which is similar to the voxel map introduced in Sec. \ref{sec::lidar-odometry}. If a successful match can be found between $\mathbb{F}_{i+1}$ and the voxel map, we obtain the relative transformation $\Delta\bT_{3,\;i+1}$ and add it as a loop closure factor to the graph. Throughout the map, we choose $m$ to be 12, and the search distance for loop closures is set to be $15$ m. In practice, we find adding loop closure factors is especially useful for eliminating the drift when conducting navigation tasks in GPS-denied regions. %We note that our proposed localization system is also compatible with other methods for loop closure detection, for example, \cite{LOOP-CLOSURE-2018} and \cite{LOOP-CLOSURE-2019}, which generate point cloud descriptors and use them for place recognition.

\begin{comment}
\section{Path Planning }
Wei, 0.5 page
\end{comment}
\section{Experiments and Results}
We performed several experiments in canals and rivers to verify the algorithms as well as demonstrate the effectiveness of the developed autonomous system. All the algorithms including SLAM, NMPC, NMHE, path planner are executed on an onboard computer (described in Section \ref{hardware}). These algorithms are updated at a rate of 10 Hz.
\subsection{SLAM Results}

\begin{figure}[ht!]
	\centering
	\subfigure[]{\includegraphics[width=.46\columnwidth]{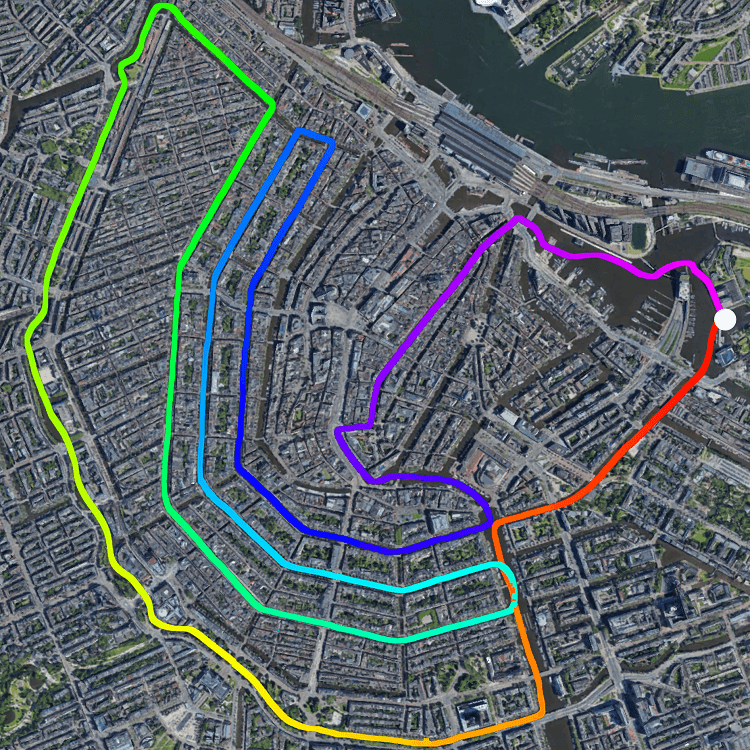}}
	\subfigure[]{\includegraphics[width=.46\columnwidth]{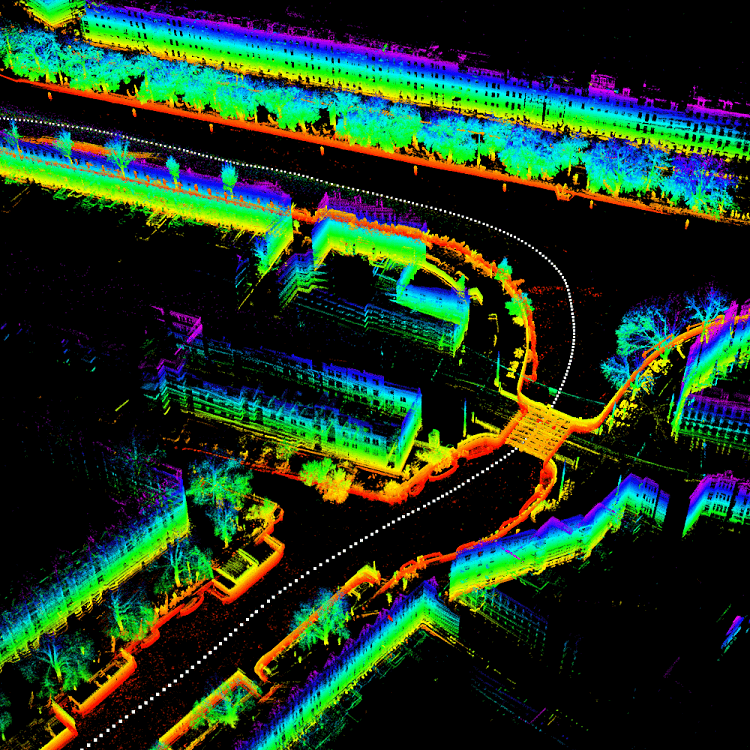}}
	\subfigure[]{\includegraphics[width=.46\columnwidth]{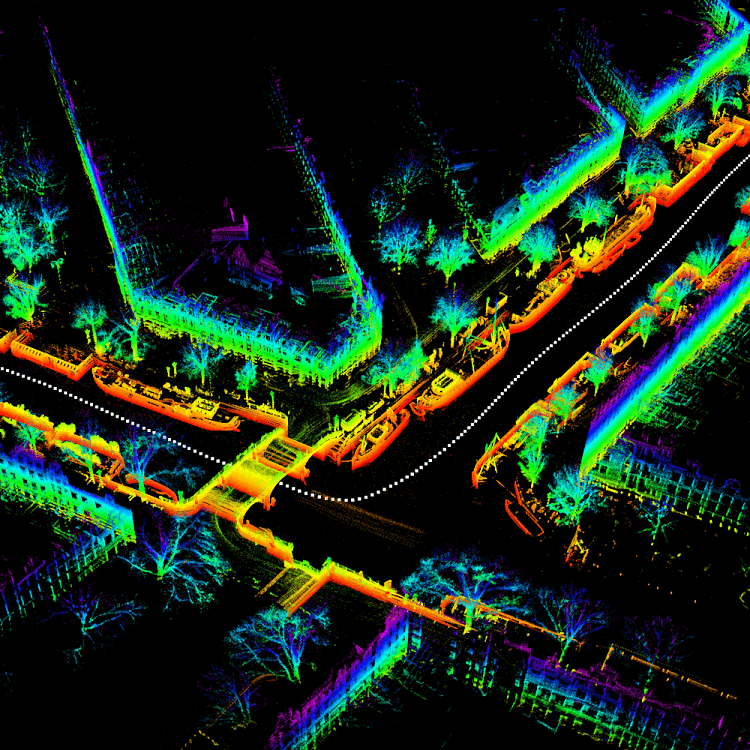}}
	\subfigure[]{\includegraphics[width=.46\columnwidth]{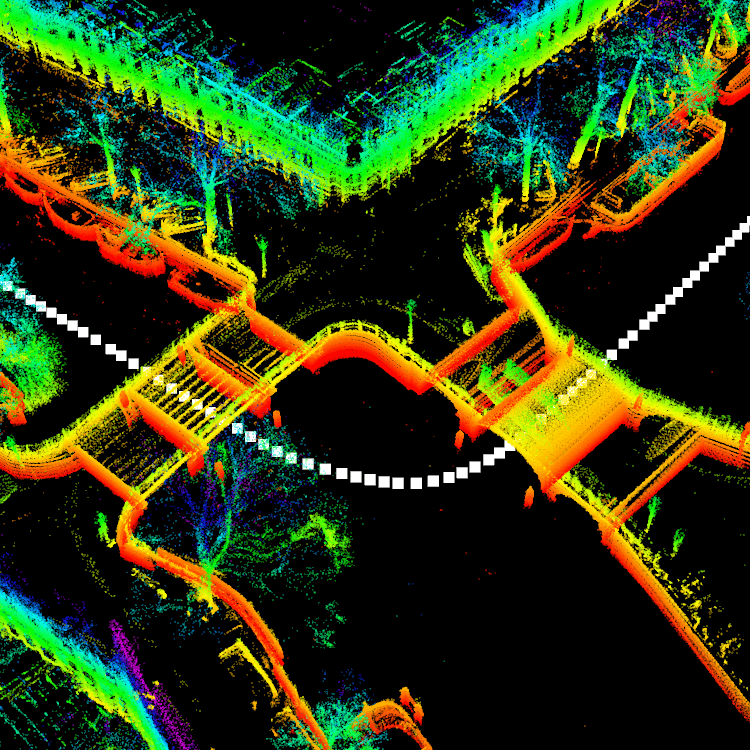}}
	\caption{The estimated trajectory (a) and representative point cloud maps (b-d) of the proposed SLAM system using a dateset gathered in Amsterdam, the Netherlands. In (a), the white dot marks the start and end location. Trajectory color variation indicates the elapse of time. The trajectory direction is clock-wise. In (b-d), color variation indicates elevation change.}
	\label{fig::amsterdam}
\end{figure}

To test the performance of the proposed SLAM system, we mount the sensor suite, which includes a Lidar, an IMU, and a GPS, on a manned boat and cruised along the canals of Amsterdam for 3 hours. We start and finish the data-gathering process at the same location. Altogether, we collected 107,656 lidar scans with an estimated trajectory length of 19,065 m. Performing SLAM in the canal environment is challenging due to several reasons. Many bridges over the canals pose degenerate scenarios, as there are few useful features when the boat is under them, similar to moving through a long, featureless corridor. Bridges and buildings obstruct the reception of GPS data and result in intermittent GPS availability throughout the dataset. We also observe numerous false directions from the lidar when direct sunlight is in the sensor field-of-view.

We test the performance of the system by disabling the insertion of GPS factor, loop closure factor, or both. When we solely use lidar odometry factors with two other factors being disabled, the system fails to produce meaningful results and suffers from great drift in a featureless environment. When we use both lidar odometry factors and GPS factors, the system can provide accurate pose estimation in the horizontal plane while suffers from drift in the vertical direction. This is because the elevation measurement from GPS is very inaccurate - giving rise to altitude errors approximating 100 m in our tests, which further motivates our usage of loop closure factors. Upon finishing processing the dataset, the proposed SLAM system achieves a relative translation error of 0.17 m when returning to the start location. We also note that the average runtime for registering a new lidar keyframe and optimizing the graph is only 79.3 ms, which is suitable for deploying the proposed system on low-powered embedded hardware.

\subsection{Results of NMPC and NMHE}
\begin{figure}[ht!]
	\centering
	\subfigure[]{\includegraphics[width=.49\columnwidth]{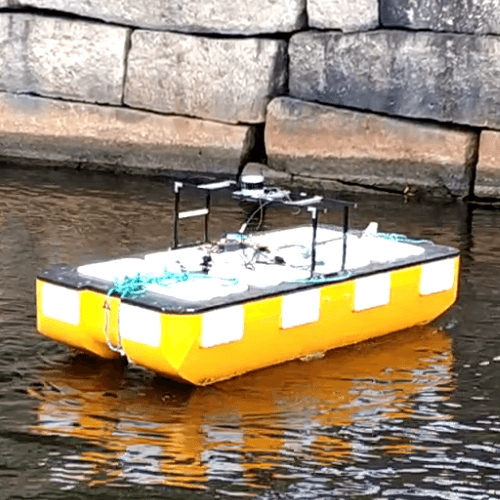}}
	\subfigure[]{\includegraphics[width=.49\columnwidth]{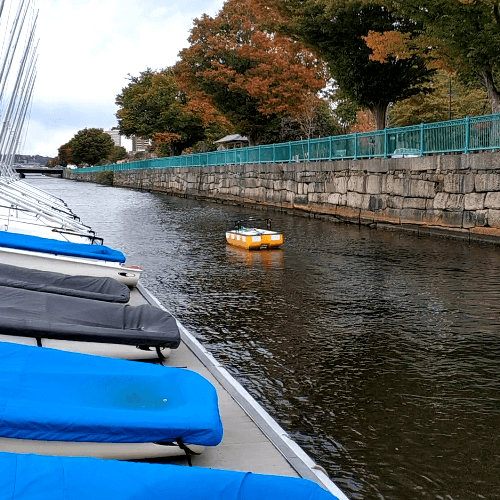}}
	\caption{Testing Roboat II in Charles River. (a) Close-up shot; (b) long shot.}
	\label{fig::charles-river}
	\vspace{-3mm}
\end{figure}

First, we experimentally estimated the dynamics model of the vessel to achieve superior performance of NMPC for tracking control.
The data was gathered when the vessel was remotely controlled to perform sinusoidal movements in Charles River (Fig. \ref{fig::charles-river}), which couples the surge, sway, and rotation motions.
The input force and the vessel velocity were recorded at a rate of 10 Hz in the experiments. The trials were repeated five times. The duration of each trial was around 150 s.  We utilized the optimization algorithm described in Section \ref{ss:dynamic_model} and identified the hydrodynamic parameters $\xi$ for Roboat II as follows: $m_{11} =172$ kg, $m_{22} =188$ kg, $m_{33} =24~ \text{kg}\cdot\text{m}^{2}$, $X_{u}=38$ kg/s, $Y_{v}=168$ kg/s and $N_{r}=16$ kg$\cdot$m$^{2}$/s.

Next, we tested the performance of the NMPC tracking and the NMHE state estimation on Roboat II in Charles River.  We use an $A^{\star}$ planner to generate the reference paths for NMPC.  The experiment was conducted as follows. First, we assigned a goal point for the planner using a ROS graphical interface, RVIZ. The planner will generate a feasible path starting from the current position and ending with the goal position of the vessel. Then, Roboat II starts to track the reference trajectories to approach the destination. After Roboat II reached the destination, we selected another goal for Roboat II which initiates a new tracking section. We repeated the same assignments several times in the experiments. NMHE was running at the same time to estimate the states of Roboat II during the experiments.

\begin{figure}[htb]
    \centering
    \includegraphics[width=0.96\linewidth] {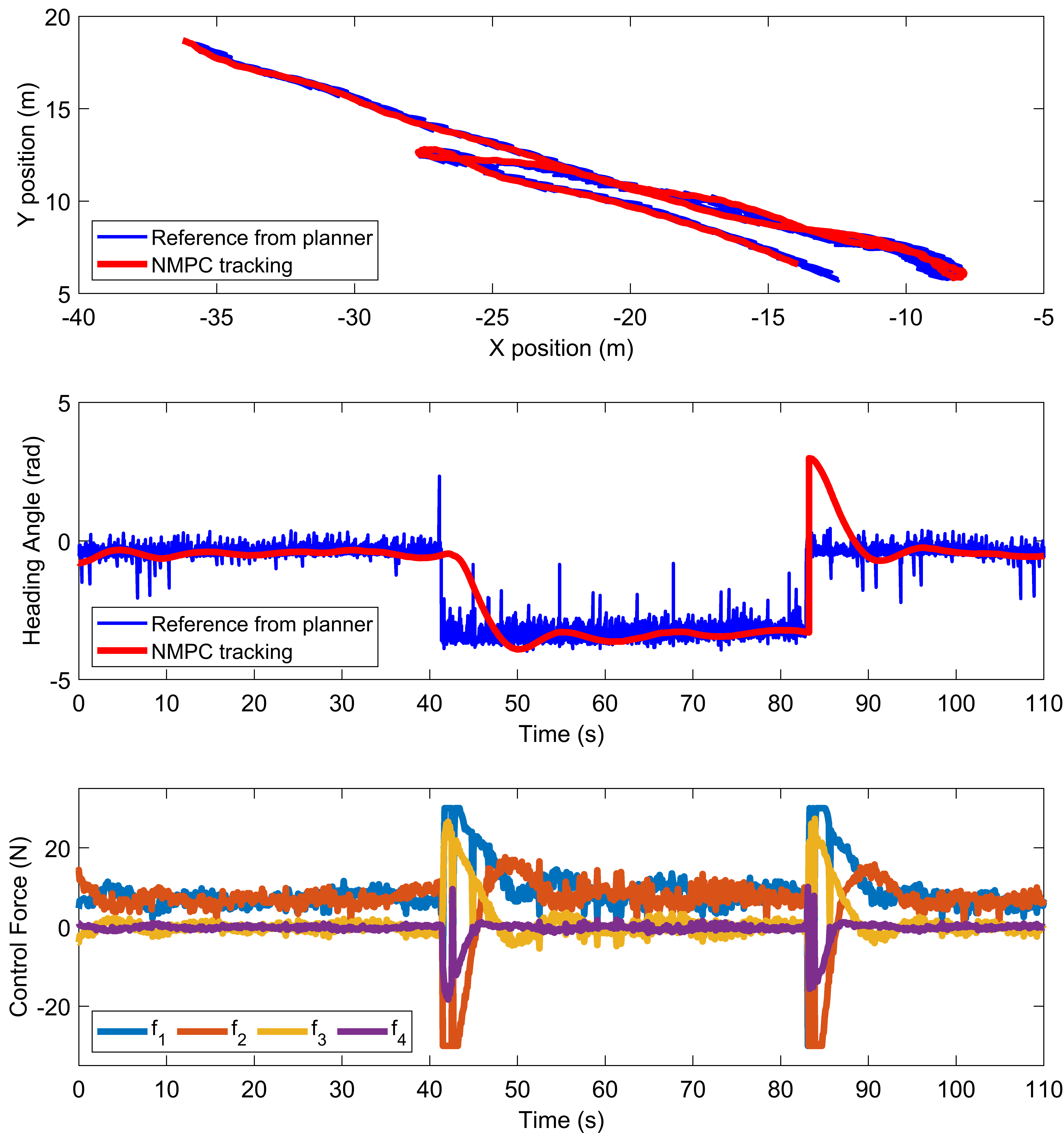}
    \caption{Performance of the NMPC tracking in the river test. (a) trajectory; (b)heading angle; (c) forces. }
    \label{MPCTracking}
    	\vspace{-3mm}
\end{figure}

Trajectory and heading angle tracking performances of Roboat II are shown in Fig. \ref{MPCTracking}.
This demonstrates that the MPC can successfully track on the references even with environmental noises in the river.  The position tracking RMSE (Root Mean Square Error) value for NMPC on Roboat II is 0.096 m while the heading angle tracking RMSE value for NMPC is 0.149 rad.
Note that we penalize (the weight coefficient) less for the heading angle than that of the positions in (\ref{OCPequation}). The reason is that we focus more on tracking the desired path accurately. Moreover, less penalization on the heading angle avoids oscillations around the desired path. The simple planner generates the whole reference path to the goal position at the rate of 10 Hz. In Fig. \ref{MPCTracking}(a) and (b), we only show 10 reference points starting from the current position of Roboat II because the generated reference path is always slightly oscillating. A more careful inspection from Fig. \ref{MPCTracking}(b) indicates that the noisy reference heading angle at each time instant during the experiments.

The control forces for Roboat II are shown in Fig.\ref{MPCTracking}(a). These generated control signals by the NMPC is restrained within the lower and upper bounds specified in (\ref{forcelimits}) in Section \ref{NMPCsection}. It is clear that the left and right thrusters contribute significantly as the system in on-track. If the system was not on-track,  all the four thrusters would contribute significantly (at around 42 s and 84 s) to help the system rapidly reach the reference positions and orientations.

\begin{figure}[htb]
    \centering
    \includegraphics[width=1.0\linewidth] {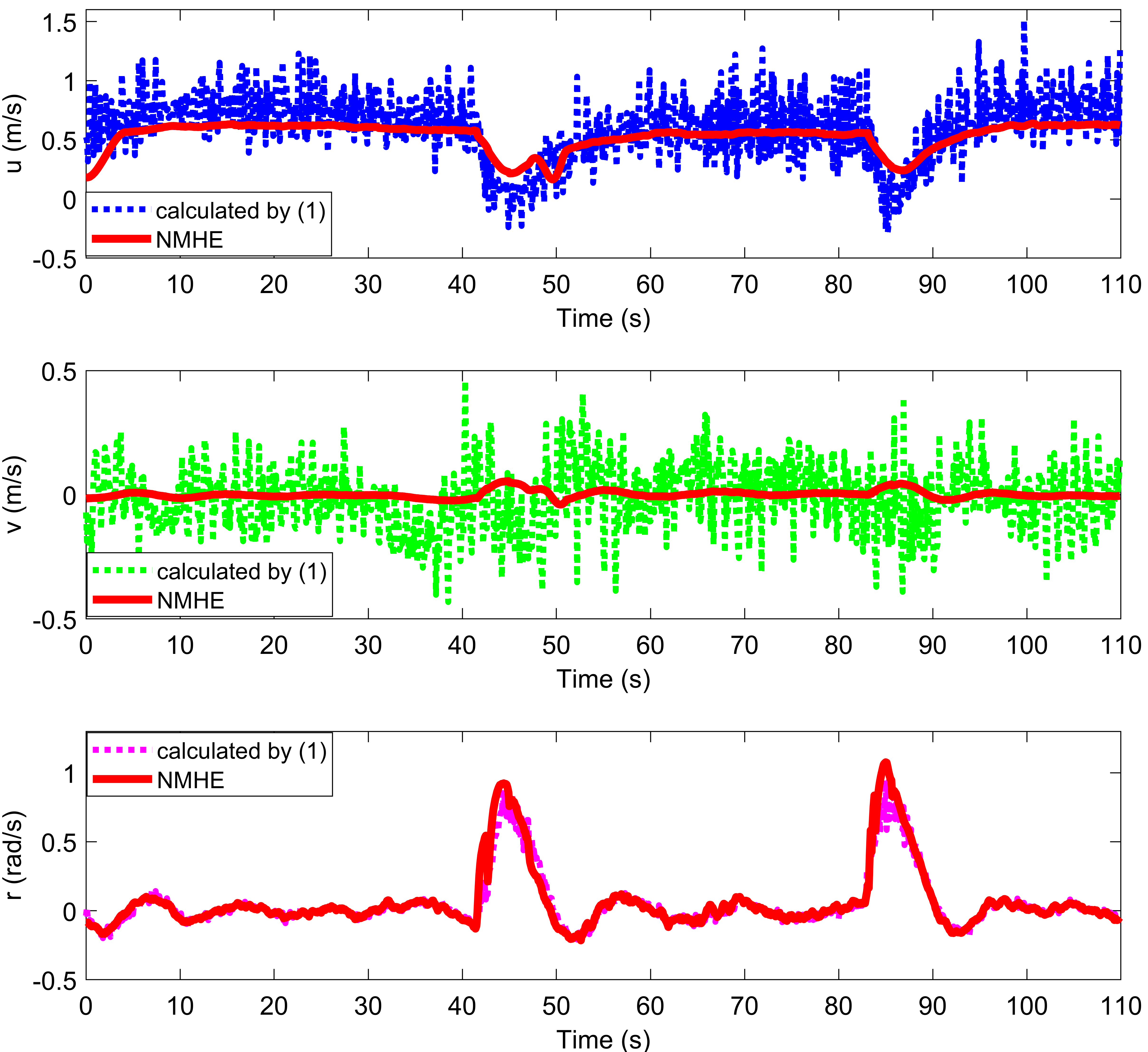}
    \caption{Performance of the NMHE state estimation in the river test. (a) linear speed $u$; (b)linear speed $v$; (c)angular speed $r$. }
    \label{NMHEvelocity}
    	\vspace{-3mm}
\end{figure}

The indirectly measured and estimated linear and angular velocities for Roboat II are shown in Fig. \ref{NMHEvelocity}.
This demonstrates that the MHE can successfully deal with noises on the measurements.
The reference for linear longitudinal velocity $u$ and lateral velocity $v$ are respectively set to 0.6 m/s and 0 m/s throughout the reference path generation. The estimates always track around these reference values as the system is on-track, which also indicates the effectiveness of the NMPC tracking. The peak and valley areas in the velocities suggest the process that Roboat II approaches the current goal position and heads to the next goal by fast turning itself.

\section{Conclusion and Future Work}
In this paper, we have developed a novel autonomous surface vessel (ASV), called Roboat II, for urban transportation. Roboat II is capable of performing accurate simultaneous localization and mapping (SLAM), receding horizon tracking control and estimation, and path planning in the urban waterways.

Our work will be extended in the following directions in the near future. First, we will explore more efficient planning algorithms to enable the vessel to handle complicated scenarios in the waterways. Second, we will apply active object detection and identification to improve Roboat's understanding of its environment for robust navigation. Third, we will estimate disturbances such as currents and waves to further improve the tracking performance in more noisy waters. Fourth, we will develop algorithms for multi-robot formation control and self-assembly on the water, enabling the construction of on-demand large-scale infrastructure.
%\section*{Acknowledgment}%
%The authors would like to thank B. Cai, Y. Zhou, X. Li, W. Tu for help in carrying out the experiments, H. Darweesh and Autoware for software consultation and advice, R. Kelly for writing revision,  and P. Leoni for contributions in building the boat hull.
%Finally, this work was supported by a grant from the Amsterdam Institute for Advanced Metropolitan Solutions (AMS) in the Netherlands.  The authors are grateful for the support.
\bibliographystyle{IEEEtran}

\bibliography{HalfScaleRoboatAutonomyIROS2020}
\end{document}